\documentclass[11pt]{article}
\usepackage[final]{acl}

\usepackage{times}
\usepackage{latexsym}
\usepackage{booktabs}
\usepackage{amsmath}
\usepackage{graphicx}
\usepackage[T1]{fontenc}
\usepackage[utf8]{inputenc}
\usepackage{microtype}
\IfFileExists{inconsolata.sty}{\usepackage{inconsolata}}{}

\title{Itgan at NADI 2026 shared task: Parameter-Efficient Whisper Adaptation for
Robust, Mixed-Dialect and Code-Switched Arabic ASR}

\author{Ibrahim Almajai \\
  Independent Researcher \\
  \texttt{i.almajai@gmail.com} \\}

\begin{document}
\maketitle

\begin{abstract}
We describe the Itgan systems for the three ASR subtasks of NADI 2026, namely robust country-level
ASR (1.1), mixed-dialect ASR (1.2), and Tunisian code-switched ASR (1.3). All three share one
recipe, Whisper adapted with LoRA on consumer GPUs, and each was carried by a different
addition to it. On 1.1, where the dialect label is given at test time, per-dialect specialists continued from a pooled adapter gave the largest gain, and the submitted system reached 57.1\% country-average WER. A post-evaluation linear probe on frozen encoder features routes utterances without the label and recovers 44\% of what oracle routing gives. On 1.2 the choice of base model mattered more than adapter
capacity, and system combination helped only once we added a decorrelated member, reaching 46.7\% WER. On 1.3 our system placed second at 14.49\% WER with the lowest CER among the leading submissions, 5.38\%. Its last 0.60 WER points came without further training, mostly from an exact weight-space average of independently trained runs, with ROVER voting adding the remainder. Every comparison
carries a paired-bootstrap test, and we report eight directions that did not work.
\end{abstract}

\section{Introduction}

The NADI 2026 shared task \citep{Sullivan-etal-2026-nadi} continues the multidialectal Arabic
speech processing track introduced the previous year \citep{talafha-etal-2025-nadi}. We
entered its three ASR subtasks: robust country-level ASR on noisy band-limited speech in eight
national dialects, scored as an unweighted mean of per-country WER (\textbf{1.1}); mixed-dialect
ASR without dialect labels (\textbf{1.2}); and Tunisian Arabic with frequent switching into
French and English (\textbf{1.3}). Section~\ref{sec:data} describes the data.

A single parameter-efficient recipe (Section~\ref{sec:common}) underlies all three systems, which
differ only in what had to be added on top. Our central finding is that the three subtasks reward
different interventions, and that identifying which one applies is worth more than additional
training. Per-dialect specialists carried 1.1, base-model choice carried 1.2 and training-free combination carried 1.3. On 1.2 the base model was worth 3.76 points while widening the adapter was a non-significant 1.02, and training-free combination was worth 0.60 points on 1.3 but nothing on 1.2 until we decorrelated the ensemble.

None of the components we use is new. LoRA, SpecAugment, weight averaging and ROVER are all
established, and our contribution is in the measurements and the system design. We give a
per-dialect specialization recipe that continues training from a pooled multi-dialect adapter,
and a linear probe on frozen encoder features that routes to those specialists without the gold
dialect label. The probe recovers 44\% of the oracle gain, with the identifier rather than the
specialists as the limit (Section~\ref{sec:t11}, Appendix~\ref{app:routing}). We show that
base-model choice, not adapter capacity, is the binding constraint for dialectal word-form
errors on 1.2, with a matched-capacity control. We argue that decoder depth is the likely reason for that advantage without isolating it from \texttt{turbo}'s post-pruning fine-tuning (Section~\ref{sec:t12}). We give exact weight-space averaging across \emph{independently
trained} runs, which beats averaging within one run even when the runs differ only in seed
(Section~\ref{sec:soup}). We also give evidence that ROVER \citep{fiscus1997post} gains come from decorrelation rather than system count (Section~\ref{sec:t12}). Every
comparison carries a paired-bootstrap test (Appendix~\ref{app:sig}), and we report eight
directions we rejected (Appendix~\ref{app:neg}).

\section{Data}
\label{sec:data}

\paragraph{Subtask 1.1.} Eight country configurations with 1{,}600 training utterances each, a
10{,}808-utterance development set holding 726 to 1{,}600 per country, and a
4{,}000-utterance blind test set split evenly. The audio is sharply band-limited, about
2.1\,kHz effective bandwidth and under 0.01\% of its energy above 4\,kHz, and noise varies far more
within a country than between countries (Appendix~\ref{app:acoustic}). Adapter selection used
the full development set, the cross-backbone comparison a 200-utterance-per-country subset.

\paragraph{Subtask 1.2.} \emph{Only} a development split of 3{,}152 utterances from more than
fifteen sub-regional varieties, without dialect labels, and a blind test set. It is the only
in-domain data for the task (Section~\ref{sec:t12}).

\paragraph{Subtask 1.3.} 38 hours of Tunisian training speech from the TEDxTN
\citep{bougares-etal-2025-tedxtn} and TuniFra \citep{choux-etal-2025-tunifra} corpora, as
24{,}893 training, 731 validation and 842 practice-test utterances, with a 2{,}123-utterance
blind test set. We drop four training utterances over 30\,s and two with empty cleaned
targets, keeping 24{,}887. Training targets and all scoring use the organizers'
\texttt{clean\_transcription} normalization, so the model is optimized for the text form the
metric scores. Code-switching is pervasive, with 33.3\% of reference words in the practice
test split in Latin script.

\section{Common Setup}
\label{sec:common}

Every system fine-tunes a Whisper checkpoint \citep{radford2023robust} with LoRA adapters
\citep{hu2022lora,peft} of rank $r=32$ and $\alpha=64$ applied to the attention projections \texttt{q\_proj}, \texttt{k\_proj}, \texttt{v\_proj}, \texttt{out\_proj} and the feed-forward layers \texttt{fc1}, \texttt{fc2} in every encoder and decoder block. We call this target set \emph{wide}, against a \emph{narrow} set of
\texttt{q\_proj} and \texttt{v\_proj} only. The frozen base is held in bfloat16 while adapter
parameters are kept in float32, which makes even a 1.54B-parameter checkpoint trainable within 12\,GB without quantization. All training and decoding ran on two consumer GPUs, an NVIDIA RTX
3080 (10\,GB) and an RTX 3060 (12\,GB). We used \texttt{transformers}
\citep{wolf2020transformers} 4.49, \texttt{peft} 0.14 and \texttt{datasets} 3.6. We report WER
and CER computed with \texttt{jiwer} 4.0.0 \citep{jiwer}, and use a paired bootstrap over utterances \citep{bisani2004bootstrap} with 4{,}000 resamples. Throughout, $P$ is the fraction of resamples in which the second system of a comparison has the lower WER, not a $p$-value, so 100\% is decisive and 50\% is chance. WER and CER are percentages throughout. Differences
between systems are absolute, in WER points, and every significance test is on an absolute
difference. Where we instead give a relative WER reduction, such as the share of a dialect's
error a specialist removes, we say so. A percentage of a gain, such as the share of the oracle
routing gain a classifier recovers, is a ratio of two absolute differences. Decoding is not matched across subtasks.
Subtask 1.1 is greedy and 1.3 beam 5, both through \texttt{transformers}, while 1.2 is beam 5
through faster-whisper \citep{fasterwhisper}. Beam search on 1.1 went untested until after the evaluation
(Appendix~\ref{app:posteval}).

\paragraph{Resources used.} Every model is a publicly released Whisper checkpoint, either
\texttt{whisper-large-v3} (1.54B parameters) or the smaller \texttt{whisper-large-v3-turbo}
(0.81B), so all three submissions are well under the shared task's 5B model-size threshold. We shorten these to \texttt{large-v3} and \texttt{turbo}. We used
\emph{no data external to NADI 2026}. Every system was trained only on its own subtask's data. The single exception is one of the five Subtask 1.2 ensemble members, trained on 1.2 data pooled with 1.1 audio (Section~\ref{sec:t12}).

\section{Subtask 1.1: Robust Country-Level ASR}
\label{sec:t11}

\paragraph{Pooled adapter.} We first train one \texttt{whisper-large-v3-turbo} adapter jointly
on all eight countries for six epochs (27.9M trainable parameters), reaching 57.6\% country-average development WER against the organizers' zero-shot baseline of 80\%.

\paragraph{Per-dialect specialists.} For four of the eight dialects we continue training that pooled adapter on the country's own 1{,}600 utterances for three epochs at learning rate $5\times10^{-5}$. The specialist keeps the shared
multi-dialect representation instead of competing with it, and a step-matched pooled control
shows the gain is specialization rather than the extra epochs (Appendix~\ref{app:whatspec}). Because the
country is known at test time, routing is deterministic and needs no dialect classifier.

\paragraph{A second backbone and a length guard.} For the remaining four countries the
transcripts come from a wide LoRA fine-tune of full \texttt{whisper-large-v3}, 57.7M trainable,
chosen per country on a matched 200-per-country subset. Its best checkpoint is epoch 4 of a six-epoch schedule. Finally
we cap any hypothesis longer than six words per second of audio. This
affects about 12 of 10{,}808 development utterances but is worth 1.0 country-average WER point, because
runaway outputs concentrate in the weakest country and every country counts equally. It also
turns out to be what makes beam search safe (Appendix~\ref{app:posteval}).

\begin{table}[t]
\centering
\small
\begin{tabular}{llcc}
\toprule
Country & Backbone & Dev $\Delta$ & Test WER \\
\midrule
Mauritania & \texttt{turbo} + spec. & $-6.5$ & 77 \\
Algeria    & \texttt{turbo} + spec. & $-3.1$ & 72 \\
Morocco    & \texttt{turbo} + spec. & $-2.9$ & 60 \\
Egypt      & \texttt{turbo} + spec. & $-1.9$ & 58 \\
Yemen      & \texttt{large-v3} & $-2.8^{*}$ & 63 \\
Jordan     & \texttt{large-v3} & $-2.7^{*}$ & 44 \\
UAE        & \texttt{large-v3} & $-4.5^{*}$ & 44 \\
Palestine  & \texttt{large-v3} & $-1.5^{*}$ & 39 \\
\midrule
\multicolumn{2}{l}{Country average} & & \textbf{57.1} \\
\bottomrule
\end{tabular}
\caption{Subtask 1.1 submitted system, all figures in \%. \emph{Dev $\Delta$} is the
full-development WER change in WER points from the pooled adapter to that country's specialist; starred rows give the change from \texttt{turbo} to
\texttt{large-v3} on the subset that decided the backbone. Every specialist gain is
significant but Jordan's (Appendix~\ref{app:sig}). Deltas are scored without the length
guard; the guarded progression of Table~\ref{tab:beam} is smaller on Mauritania and Yemen,
where the guard does part of the specialist's work (Appendix~\ref{app:whatspec}).}
\label{tab:t11}
\end{table}

\paragraph{Results.} The system scores 57.1\% country-average WER / 28\% CER on the blind test set, seventh of nine teams, against 43.2\% for the leader. We cannot say what separates them
from us; our systems saw only this subtask's 12{,}800 training utterances.
Table~\ref{tab:t11} shows the specialist gain tracking each dialect's headroom almost
monotonically. Without the length guard, as Table~\ref{tab:t11} is scored, a specialist removes
4 to 8\% of that dialect's remaining error. Under the guard it removes 3.6 to 4.5\%, because
the guard overlaps with the specialist on the over-generating dialects
(Appendix~\ref{app:whatspec}). After the evaluation closed we completed the set of eight specialists, and found beam search with the guard worth a further 1.1 points on top of all eight,
reaching 53.55\% in \emph{development} WER (Appendix~\ref{app:posteval}).

\section{Subtask 1.2: Mixed-Dialect ASR}
\label{sec:t12}

We partition the 3{,}152-utterance development split (Section~\ref{sec:data}) by SHA-1 hash of the
utterance id into 2{,}537 training and 615 held-out utterances, make every design decision on
the held-out portion, then retrain the submitted model on all 3{,}152. Retraining on everything was worth 1.0 WER point and changed 86\% of test transcripts. It also spent the local
evaluation, since a model that has seen every held-out id can no longer be scored on it.

\begin{table}[t]
\centering
\small
\setlength{\tabcolsep}{4pt}
\begin{tabular}{lrr}
\toprule
System & Trainable & WER \\
\midrule
\texttt{turbo}, zero-shot          & n/a   & 60.04 \\
\texttt{turbo} + LoRA (narrow)     & 6.6M  & 56.54 \\
\texttt{turbo} + LoRA (wide)       & 27.9M & 55.51 \\
\texttt{large-v3}, zero-shot (int8) & n/a & 57.21 \\
\texttt{large-v3} + LoRA (wide, $r$\,16) & 28.8M & 52.88 \\
\texttt{large-v3} + LoRA (wide) & 57.7M & \textbf{51.76} \\
\bottomrule
\end{tabular}
\caption{Subtask 1.2, 615 held-out utterances (\%). Differences quoted in the text are
paired-bootstrap estimates (Appendix~\ref{app:sig}) and can differ by about 0.01 from
differences of the values printed here.}
\label{tab:t12}
\end{table}

\paragraph{The base model is the lever.} Table~\ref{tab:t12} isolates the effect of the base
model. Switching from \texttt{turbo} to full \texttt{large-v3} was worth 3.76 WER
points, against 4.52 points for all of \texttt{turbo}'s fine-tuning. The decisive comparison is the fourth row
against the third, where \emph{untrained} \texttt{large-v3}, quantized to int8, nearly matches
a fully fine-tuned \texttt{turbo}. Adapter size explains only part of the gap. The fifth row halves the
\texttt{large-v3} adapter's rank to match \texttt{turbo}'s 27.9M trainable parameters, and
\texttt{large-v3} still leads \texttt{turbo} by 2.65 points ($P=100\%$, 95\% interval
$[-3.85,-1.39]$); the extra rank is worth the remaining 1.11 points ($P=99\%$). Decoder depth is
the most plausible reason for what is left. \texttt{turbo} has 4 decoder layers against 32 in its encoder, whereas full \texttt{large-v3} has 32 of each.
This task's dominant error is producing the correct dialectal word form, which is decoder
work. The comparison does not isolate depth, however. The \texttt{turbo} checkpoint is \texttt{large-v3} with the decoder pruned to 4 layers and then trained further, and that training also moved its encoder
weights, which we verified against the released checkpoints. The two rows therefore differ in
depth, in that extra training, and in int8 quantization on the zero-shot row, which if
anything handicaps it. Separating depth from \texttt{turbo}'s extra training would take a pruned
\texttt{large-v3} without that training, which we did not run. The depth reading
does account for an otherwise puzzling asymmetry. Widening the adapter target set was worth
roughly 8 WER points on Subtask 1.1, whose wider run also trained twice as long, but only 1.02
on a schedule-matched pair here. Whatever the mechanism, adapter capacity did not substitute
for the base-model change.

\paragraph{Decorrelation beats system count.} Our submitted entry combines five systems by word-level ROVER voting, but the path there is the informative part. Our first
ensemble combined two \texttt{large-v3} fine-tunes with a \texttt{turbo} one and scored 48\%, \emph{identical to its own pivot} (the member whose hypothesis the vote defaults to), despite an offline gain that had promised $-0.65$ points. A fourth
member, trained on 1.2 data pooled with 12{,}800 utterances of Subtask 1.1 audio at roughly
37\% in-domain, changed that. It disagrees with the \texttt{large-v3} fine-tune we can score
offline on 91\% of held-out lines, while scoring within 0.33 points of it. The four-system ensemble scored 46.7\%. A fifth member, zero-shot \texttt{large-v3}, moved nothing further, and the submitted archive is that five-system build.

\paragraph{Results.} Decoding merges the adapter, converts to CTranslate2 \citep{klein-etal-2020-opennmt}, and adds Silero VAD \citep{silerovad}
with no conditioning on previous text. The leaderboard progression was 52\% for a wide \texttt{turbo} fine-tune, 49\% for \texttt{large-v3} on the 2{,}537 partition, 48\% for \texttt{large-v3} on all 3{,}152, and 46.7\% WER / 19\% CER for the submitted ensemble, third of four teams. Roughly 5 WER points of the reported error on this subtask are unwinnable orthographic variation
(Appendix~\ref{app:noise}).

\section{Subtask 1.3: Code-Switched ASR}
\label{sec:t13}

\begin{table*}[t]
\centering
\small
\begin{tabular}{llccccc}
\toprule
& & \multicolumn{2}{c}{Validation (731)} & \multicolumn{2}{c}{Practice test (842)} & Blind test (2{,}123) \\
\cmidrule(lr){3-4}\cmidrule(lr){5-6}\cmidrule(lr){7-7}
System & Trainable & WER & CER & WER & CER & WER / CER \\
\midrule
Zero-shot \texttt{whisper-large-v3}$^{\dagger}$ & n/a & 85.00 & 59.84 & 87.57 & 64.40 & n/a \\
LoRA (\texttt{q,v}), 3k steps & 15.7M & 21.62 & 9.11 & 24.68 & 11.79 & n/a \\
LoRA (wide), 5k steps & 57.7M & 19.84 & 8.26 & 21.74 & 10.16 & n/a \\
\quad + dropout 0.15, SpecAugment & 57.7M & 19.46 & 8.29 & 21.19 & 10.08 & 15.09 / 5.56 \\
\quad + adapter soup ($N{=}4$) & n/a & 17.19 & \textbf{6.60} & \textbf{20.14} & \textbf{9.51} & 14.55 / 5.40 \\
\quad + ROVER (3 systems) & n/a & \textbf{17.09} & 6.61 & 20.15 & 9.56 & \textbf{14.49 / 5.38} \\
\bottomrule
\end{tabular}
\caption{Subtask 1.3 results (\%). $^{\dagger}$the organizers' baseline notebook; under the beam-5 decode that all trained rows use, the checkpoint loops and scores 103.2 / 117.8. The final
system reduces blind-test WER by 0.60 points over the best single trained model.}
\label{tab:main}
\end{table*}

\paragraph{Training and regularization.} We fine-tune \texttt{whisper-large-v3} with the wide
target set (57.7M trainable, 3.6\% of the model) for 5{,}000 steps
($\approx$3.2 epochs) with an effective batch of 16, peak learning rate $5\times10^{-4}$, 50
warmup steps, linear decay and gradient checkpointing. Our first wide
run memorized heavily, with training loss at 0.04--0.05 while held-out loss stagnated near
0.31. Raising LoRA dropout from 0.05 to 0.15 and enabling SpecAugment
\citep{park2019specaugment} behaved as regularization should. At matched step 3{,}000,
training loss \emph{rose} from 0.229 to 0.293 while held-out loss \emph{fell} from 0.340 to
0.322, shrinking the generalization gap by a factor of 3.9.

\subsection{Cross-run adapter souping}
\label{sec:soup}

Weight averaging of independently fine-tuned models improves accuracy
\citep{wortsman2022model,izmailov2018swa}, but applying it to LoRA requires care. The adapter's
contribution $\Delta W_i = \frac{\alpha}{r} B_i A_i$ is \emph{bilinear} in the factors, so
averaging $A_i$ and $B_i$ elementwise does not average the $\Delta W_i$.
Concatenating the factors does. For $N$ adapters of rank $r$, set
\begin{equation}
A' = \bigl[\,A_1^{\top}\ \cdots\ A_N^{\top}\,\bigr]^{\top}\!,\;
B' = \bigl[\,B_1\ \cdots\ B_N\,\bigr]\!,
\end{equation}
with $r' = Nr$, and keep $\alpha' = \alpha$. The scaling becomes
$\alpha'/r' = \frac{1}{N}\cdot\frac{\alpha}{r}$ and
\begin{equation}
\frac{\alpha'}{r'} B' A' = \frac{1}{N}\sum_{i=1}^{N}\frac{\alpha}{r}B_i A_i
= \frac{1}{N}\sum_{i=1}^{N}\Delta W_i,
\end{equation}
i.e.\ the exact mean, as an ordinary rank-$Nr$ adapter that any PEFT implementation can load
and that we merge into the base before decoding, so the extra rank is free at inference. The construction is not new, and \texttt{peft} provides it as the \texttt{cat} combination type of \texttt{add\_weighted\_adapter} \citep{peft}. We restate it because the
elementwise mistake is easy to make, and verified it on the submitted soup (maximum deviation
$9\times10^{-8}$ in float32).

Our submitted soup averages the regularized run's last three checkpoints plus the
unregularized run's final adapter. The gain is decorrelation, not ingredient count. At two ingredients, the cross-recipe pair, one checkpoint from each run, scores 18.06\%, and the same-run pair 19.28\%, a gap of 1.23 points ($P=100\%$). A second checkpoint from the same run is worth only 0.18 points, which is not significant. Two ingredients from different runs
beat three from one (18.66\%), and six ingredients, with the weaker run at half weight, score 17.81\%.
The regularized and unregularized runs differ in seed as well as in regularization, so after the evaluation we
retrained the regularized recipe at a second seed, changing nothing but initialization and
data order. The second seed alone scores 18.71\% against 19.46\%, a spread that is not significant
($P=87\%$). Averaging the two seeds scores
16.91\%, which is 2.38 points below the same-run pair ($P=100\%$), 1.16 points below the cross-recipe pair ($P=99\%$), and indistinguishable from the submitted four-ingredient soup
(17.19\%, $P=87\%$). Seed decorrelation alone accounts for the cross-run gain.

\subsection{ROVER and results}

We finally combine the soup as pivot with the two trained models as voters, by word-level majority voting over aligned token positions, with the pivot breaking the ties, which arise only where all three disagree. This modifies 63 of 2{,}123 blind utterances. Its effect is small and not uniformly positive, at $-0.10$ WER points on validation, $-0.06$ on the blind set, $+0.01$ on the practice test split, so it should be read as noise-level rather than
an established gain.

Table~\ref{tab:main} gives the progression. Fine-tuning accounts for the bulk of the
improvement; the two training-free steps supply the last 0.60 points. The submission
placed second among the six submissions, 0.08 WER points behind the first-placed system. Our CER of 5.38\%
is the lowest among the leading submissions, below the first-placed system's 5.55\%, although
the subtask is ranked on WER. Embedded French and English
words are recognized \emph{better} than the surrounding Tunisian Arabic
(Appendix~\ref{app:script}), so the bottleneck is the dialect, not the code-switching.

\section{Discussion and Future Work}

Our main lesson is to diagnose the binding constraint before spending GPU-hours. Each diagnosis was cheap: a loss gap, a
zero-shot comparison, a disagreement rate.

The Subtask 1.2 description encourages combining dialect identification with transcription.
We could not score that on 1.2, which has no dialect labels, but Subtask 1.1 lets us measure
what it is worth, since there the label is known and routing can be scored. Oracle routing to the specialists gains 1.9 country-average WER points over the pooled adapter, a 3.4\% relative reduction. A linear probe on frozen encoder features, at 65\% identification accuracy, recovers 0.8 points of that, 44\%
(Appendix~\ref{app:routing}). The shortfall is structural. A specialist costs 1.4 points on the
seven dialects it did not train on against a 1.9 point gain on its own, so misrouting costs about seven tenths of what correct routing gains, and identification accuracy sets the return. The
identifier, not the specialists, is where further effort belongs. Subtask 1.2 has more than fifteen
varieties and no labels, against eight known classes on 1.1, so the recoverable share there is
likely lower. NADI 2026 ran a separate spoken dialect identification task, which we did not
enter. On the evidence here that was the wrong call: an identifier built for that track drops
straight into this front-end, so the two are complementary rather than separate problems.

\section{Conclusion}

We described three NADI 2026 ASR systems from one recipe on consumer hardware, reaching 57.1\% country-average WER on 1.1 (seventh of nine), 46.7\% on 1.2 (third of four), 14.49\% on 1.3
(second of six). Each was carried by a different intervention: specialists,
base-model choice, cross-run averaging.

\section*{Limitations}

We made our Subtask 1.3 design decisions on a single 731-utterance development split, which
our bootstrap analysis shows cannot resolve differences below roughly 1.5 WER points. Some
reported development gains are therefore partly selection noise, as the transfer analysis in
Appendix~\ref{app:transfer} quantifies. We trained one seed per configuration throughout, so
run-to-run variance is not separated from the effect of the changes we made. Repeating the
Mauritania specialist at a second seed moves its development WER by 0.4 points, against a
guarded specialist gain of 2.8 points (Appendix~\ref{app:whatspec}). Repeating the regularized
Subtask 1.3 recipe at a second seed moves its validation WER by 0.75 points, against the 1.5 points the split can resolve. Both bound
the concern without removing it. The one comparison we re-ran across seeds, cross-run averaging,
survived (Section~\ref{sec:soup}). For Subtask 1.2
we trained the submitted model on the entire development split, so it cannot be scored
offline at all and its blind-test result is the only measurement of it we have. We chose the
Subtask 1.1 per-country backbone on a 200-utterance-per-country subset whose margin was, for three of the four countries it decided, under 3 WER points. We did not tune
decoding hyperparameters beyond the ablation reported for Subtask 1.2, and Subtask 1.1 was
decoded greedily throughout, a cost we later measured at 1.1 to 1.3 country-average WER points
(Appendix~\ref{app:posteval}). Finally, our Subtask
1.3 error analysis uses script as a proxy for language, which counts French and English words
written in Arabic script as Arabic.

\newpage
\bibliography{refs}  

\begin{thebibliography}{19}
\providecommand{\natexlab}[1]{#1}

\bibitem[{Bisani and Ney(2004)}]{bisani2004bootstrap}
Maximilian Bisani and Hermann Ney. 2004.
\newblock \href {https://doi.org/10.1109/ICASSP.2004.1326009} {Bootstrap
  estimates for confidence intervals in {ASR} performance evaluation}.
\newblock In \emph{2004 IEEE International Conference on Acoustics, Speech, and
  Signal Processing}, volume~1, pages I--409--I--412. IEEE.

\bibitem[{Bougares et~al.(2025)Bougares, Mdhaffar, Elleuch, and
  Est{\`e}ve}]{bougares-etal-2025-tedxtn}
Fethi Bougares, Salima Mdhaffar, Haroun Elleuch, and Yannick Est{\`e}ve. 2025.
\newblock \href {https://doi.org/10.18653/v1/2025.arabicnlp-main.22}
  {{TED}x{TN}: A three-way speech translation corpus for code-switched
  {T}unisian {A}rabic - {E}nglish}.
\newblock In \emph{Proceedings of The Third Arabic Natural Language Processing
  Conference}, pages 278--287, Suzhou, China. Association for Computational
  Linguistics.

\bibitem[{Cao et~al.(2022)Cao, Abdulatif, and Yang}]{cao2022cmgan}
Ruizhe Cao, Sherif Abdulatif, and Bin Yang. 2022.
\newblock \href {https://doi.org/10.21437/Interspeech.2022-517} {{CMGAN}:
  Conformer-based metric {GAN} for speech enhancement}.
\newblock In \emph{{Interspeech 2022}}, pages 936--940.

\bibitem[{Choux et~al.(2025)Choux, Avila, Crego, Bougares, and
  Laurent}]{choux-etal-2025-tunifra}
Alex Choux, Marko Avila, Josep Crego, Fethi Bougares, and Antoine Laurent.
  2025.
\newblock \href {https://doi.org/10.18653/v1/2025.arabicnlp-main.5}
  {{T}uni{F}ra: A {T}unisian {A}rabic speech corpus with orthographic
  transcriptions and {F}rench translations}.
\newblock In \emph{Proceedings of The Third Arabic Natural Language Processing
  Conference}, pages 64--68, Suzhou, China. Association for Computational
  Linguistics.

\bibitem[{Fiscus(1997)}]{fiscus1997post}
Jonathan~G. Fiscus. 1997.
\newblock \href {https://doi.org/10.1109/ASRU.1997.659110} {A post-processing
  system to yield reduced word error rates: Recognizer output voting error
  reduction ({ROVER})}.
\newblock In \emph{1997 IEEE workshop on automatic speech recognition and
  understanding proceedings}, pages 347--354. IEEE.

\bibitem[{Hu et~al.(2022)Hu, Shen, Wallis, Allen-Zhu, Li, Wang, Wang, and
  Chen}]{hu2022lora}
Edward~J. Hu, Yelong Shen, Phillip Wallis, Zeyuan Allen-Zhu, Yuanzhi Li, Shean
  Wang, Lu~Wang, and Weizhu Chen. 2022.
\newblock \href {https://arxiv.org/abs/2106.09685} {{LoRA}: Low-rank adaptation
  of large language models}.
\newblock In \emph{International Conference on Learning Representations
  (ICLR)}.

\bibitem[{Izmailov et~al.(2018)Izmailov, Podoprikhin, Garipov, Vetrov, and
  Wilson}]{izmailov2018swa}
Pavel Izmailov, Dmitrii Podoprikhin, Timur Garipov, Dmitry Vetrov, and
  Andrew~Gordon Wilson. 2018.
\newblock \href {http://auai.org/uai2018/proceedings/papers/313.pdf} {Averaging
  weights leads to wider optima and better generalization}.
\newblock In \emph{Conference on Uncertainty in Artificial Intelligence (UAI)},
  pages 876--885.

\bibitem[{Klein and Ashraf(2023)}]{fasterwhisper}
Guillaume Klein and Mahmoud Ashraf. 2023.
\newblock faster-whisper: Faster {W}hisper transcription with {CTranslate2}.
\newblock \url{https://github.com/SYSTRAN/faster-whisper}.

\bibitem[{Klein et~al.(2020)Klein, Hernandez, Nguyen, and
  Senellart}]{klein-etal-2020-opennmt}
Guillaume Klein, Fran{\c{c}}ois Hernandez, Vincent Nguyen, and Jean Senellart.
  2020.
\newblock \href {https://aclanthology.org/2020.amta-research.9/} {The
  {O}pen{NMT} neural machine translation toolkit: 2020 edition}.
\newblock In \emph{Proceedings of the 14th Conference of the Association for
  Machine Translation in the Americas (Volume 1: Research Track)}, pages
  102--109, Virtual. Association for Machine Translation in the Americas.

\bibitem[{Liu et~al.(2021)Liu, Kong, Tian, Zhao, Wang, Huang, and
  Wang}]{liu2021voicefixer}
Haohe Liu, Qiuqiang Kong, Qiao Tian, Yan Zhao, DeLiang Wang, Chuanzeng Huang,
  and Yuxuan Wang. 2021.
\newblock \href {https://arxiv.org/abs/2109.13731} {{VoiceFixer}: Toward
  general speech restoration with neural vocoder}.
\newblock \emph{arXiv preprint arXiv:2109.13731}.

\bibitem[{Mangrulkar et~al.(2022)Mangrulkar, Gugger, Debut, Belkada, Paul,
  Bossan, and Tietz}]{peft}
Sourab Mangrulkar, Sylvain Gugger, Lysandre Debut, Younes Belkada, Sayak Paul,
  Benjamin Bossan, and Marian Tietz. 2022.
\newblock {PEFT}: State-of-the-art parameter-efficient fine-tuning methods.
\newblock \url{https://github.com/huggingface/peft}.

\bibitem[{Park et~al.(2019)Park, Chan, Zhang, Chiu, Zoph, Cubuk, and
  Le}]{park2019specaugment}
Daniel~S. Park, William Chan, Yu~Zhang, Chung-Cheng Chiu, Barret Zoph, Ekin~D.
  Cubuk, and Quoc~V. Le. 2019.
\newblock \href {https://doi.org/10.21437/Interspeech.2019-2680}
  {{SpecAugment}: A simple data augmentation method for automatic speech
  recognition}.
\newblock In \emph{{Interspeech 2019}}, pages 2613--2617.

\bibitem[{Radford et~al.(2023)Radford, Kim, Xu, Brockman, McLeavey, and
  Sutskever}]{radford2023robust}
Alec Radford, Jong~Wook Kim, Tao Xu, Greg Brockman, Christine McLeavey, and
  Ilya Sutskever. 2023.
\newblock \href {https://proceedings.mlr.press/v202/radford23a.html} {Robust
  speech recognition via large-scale weak supervision}.
\newblock In \emph{International conference on machine learning}, pages
  28492--28518. PMLR.

\bibitem[{{Silero Team}(2024)}]{silerovad}
{Silero Team}. 2024.
\newblock Silero {VAD}: pre-trained enterprise-grade voice activity detector
  ({VAD}), number detector and language classifier.
\newblock \url{https://github.com/snakers4/silero-vad}.

\bibitem[{Sullivan et~al.(2026)Sullivan, Talafha, Ashraf, Bougares, Elleuch,
  Zhang, Elmadany, Mohamed, Mdhaffar, Est{\`e}ve, Elhoseiny, Luqman, Habash,
  and Abdul-Mageed}]{Sullivan-etal-2026-nadi}
Peter Sullivan, Bashar Talafha, Ahmed Ashraf, Fethi Bougares, Haroun Elleuch,
  Chiyu Zhang, AbdelRahim Elmadany, Youssef Mohamed, Salima Mdhaffar, Yannick
  Est{\`e}ve, Mohamed Elhoseiny, Hamzah Luqman, Nizar Habash, and Muhammad
  Abdul-Mageed. 2026.
\newblock {NADI}-2026: The second multidialectal {A}rabic speech processing
  shared task.
\newblock In \emph{Proceedings of the Fourth Arabic Natural Language Processing
  Conference (ArabicNLP 2026)}, Budapest, Hungary. Association for
  Computational Linguistics.

\bibitem[{Talafha et~al.(2025)Talafha, Toyin, Sullivan, Elmadany, Juma,
  Djanibekov, Zhang, Alshehhi, Aldarmaki, Jarrar, Habash, and
  Abdul-Mageed}]{talafha-etal-2025-nadi}
Bashar Talafha, Hawau~Olamide Toyin, Peter Sullivan, AbdelRahim~A. Elmadany,
  Abdurrahman Juma, Amirbek Djanibekov, Chiyu Zhang, Hamad Alshehhi, Hanan
  Aldarmaki, Mustafa Jarrar, Nizar Habash, and Muhammad Abdul-Mageed. 2025.
\newblock \href {https://doi.org/10.18653/v1/2025.arabicnlp-sharedtasks.99}
  {{NADI} 2025: The first multidialectal {A}rabic speech processing shared
  task}.
\newblock In \emph{Proceedings of The Third Arabic Natural Language Processing
  Conference: Shared Tasks}, pages 720--733, Suzhou, China. Association for
  Computational Linguistics.

\bibitem[{Vaessen(2025)}]{jiwer}
Nik Vaessen. 2025.
\newblock {jiwer}: Evaluate your speech-to-text system with similarity
  measures.
\newblock \url{https://github.com/jitsi/jiwer/releases/tag/v4.0.0}.
\newblock Version 4.0.0.

\bibitem[{Wolf et~al.(2020)Wolf, Debut, Sanh, Chaumond, Delangue, Moi, Cistac,
  Rault, Louf, Funtowicz, Davison, Shleifer, von Platen, Ma, Jernite, Plu, Xu,
  Le~Scao, Gugger, Drame, Lhoest, and Rush}]{wolf2020transformers}
Thomas Wolf, Lysandre Debut, Victor Sanh, Julien Chaumond, Clement Delangue,
  Anthony Moi, Pierric Cistac, Tim Rault, Remi Louf, Morgan Funtowicz, Joe
  Davison, Sam Shleifer, Patrick von Platen, Clara Ma, Yacine Jernite, Julien
  Plu, Canwen Xu, Teven Le~Scao, Sylvain Gugger, and 3 others. 2020.
\newblock \href {https://doi.org/10.18653/v1/2020.emnlp-demos.6} {Transformers:
  State-of-the-art natural language processing}.
\newblock In \emph{Proceedings of the 2020 conference on empirical methods in
  natural language processing: system demonstrations}, pages 38--45.

\bibitem[{Wortsman et~al.(2022)Wortsman, Ilharco, Gadre, Roelofs,
  Gontijo-Lopes, Morcos, Namkoong, Farhadi, Carmon, Kornblith, and
  Schmidt}]{wortsman2022model}
Mitchell Wortsman, Gabriel Ilharco, Samir~Yitzhak Gadre, Rebecca Roelofs,
  Raphael Gontijo-Lopes, Ari~S. Morcos, Hongseok Namkoong, Ali Farhadi, Yair
  Carmon, Simon Kornblith, and Ludwig Schmidt. 2022.
\newblock \href {https://proceedings.mlr.press/v162/wortsman22a.html} {Model
  soups: Averaging weights of multiple fine-tuned models improves accuracy
  without increasing inference time}.
\newblock In \emph{International conference on machine learning}, pages
  23965--23998. PMLR.

\end{thebibliography}

\appendix

\section{Acoustic Character of Subtask 1.1}
\label{app:acoustic}

Effective bandwidth is about 2.1\,kHz, and under 0.01\% of energy lies above 4\,kHz in every
country, so 43\% of Whisper's 128 mel bins peak above that line and carry nothing. A
per-utterance SNR proxy, the ratio of the 90th to the 10th percentile of frame energy, runs
from 11.2 to 40.0\,dB between its 5th and 95th percentiles, with 14.1\% of the 10{,}808
development utterances below 15\,dB. Splitting each country at its \emph{own} median, the noisier half decodes 8.6 WER points worse in all eight countries, whereas per-country mean SNR
ranges only 21.0 to 29.9\,dB, so country means hide the effect entirely. Across countries,
dialect dominates instead: per-country WER spans 39 points. It correlates $-0.14$ with
bandwidth and $-0.52$ with mean SNR, but with $n=8$ those have 95\% intervals of
$[-0.77,+0.63]$ and $[-0.90,+0.29]$. Eight countries cannot separate either from zero, and we
report them only to show that neither acoustic axis explains the spread.

Model selection on the training-time evaluation subset is unsafe here: it understated WER by about 2.4 points and reversed the ranking of two configurations that full-development scoring
showed to be equivalent.

\section{Post-Evaluation}
\label{app:posteval}

Every number in this appendix is \emph{development} WER and none is comparable with the
blind-test figures of Table~\ref{tab:t11}. The test references were never released, so no
post-evaluation change can be scored on the metric the task was ranked by.

\paragraph{All eight specialists.} By the submission deadline six countries had specialists, the four of Table~\ref{tab:t11} at three epochs and Yemen and Jordan at a cheaper single annealed epoch, neither of which the backbone comparison kept. After the evaluation closed we added UAE and Palestine on that same one-epoch recipe. Against the pooled \texttt{turbo} adapter,
unguarded as in Table~\ref{tab:t11}, Yemen improved by 3.9 WER points, UAE by 1.6, Palestine by 1.1 and Jordan by 0.4, following the headroom order of Table~\ref{tab:t11} except for
Jordan. All eight dialects are therefore reachable in under 3
GPU-hours on the one-epoch recipe, against the 4.6 hours the four three-epoch specialists cost
on their own. These four gains are measured against the pooled \texttt{turbo} adapter, not against the
\texttt{large-v3} transcripts the submission actually used for these countries, so they do not
translate into a submission-time improvement.

\paragraph{Beam search, and why the length guard is what makes it usable.} Every Subtask 1.1
number above was decoded greedily. Re-decoding the full development set with beam 5 at batch 8
costs about 2.3 times the greedy wall clock (Egypt's 1{,}600 utterances: 306 and 348\,s greedy
against 754\,s twice, on an idle RTX 3060, model load included) and is worth 1.3 country-average WER points on the pooled adapter. It is not safe on its own: on the two dialects that over-generate it
makes things \emph{worse}, taking Mauritania from 73.3 to 77.1 and Algeria from 66.6 to 68.5 relative to their greedy specialists. Beam search without a length penalty prefers longer
hypotheses, and the number of runaway outputs the guard has to cut on the specialists rises from 8 to 26
across the corpus, 1 to 15 on Mauritania alone. On the pooled adapter it rises from 14 to 35. Applying the guard removes the interaction and beam
search then improves \emph{all eight} countries, Mauritania included (Table~\ref{tab:beam}).

\begin{table}[h]
\centering
\small
\setlength{\tabcolsep}{4pt}
\begin{tabular}{lc}
\toprule
Configuration & Country-average WER (\%) \\
\midrule
pooled adapter, greedy            & 57.59 \\
\quad + length guard             & 56.58 \\
\quad + beam 5                   & 55.31 \\
all eight specialists + guard     & 54.67 \\
\quad + beam 5                   & \textbf{53.55} \\
\bottomrule
\end{tabular}
\caption{Post-evaluation development results for Subtask 1.1. The length guard applies from
the second row down, unlike the per-country deltas of Table~\ref{tab:t11}, which are unguarded.
The only training beyond Table~\ref{tab:t11} is the four one-epoch specialists described above;
the guard and the beam are free.}
\label{tab:beam}
\end{table}

Per country, adding beam to the guarded specialists helps everywhere, from 0.3 points on Mauritania to 1.7 on Morocco, at $P=100\%$ on five of the eight and 98\% on Yemen. The two exceptions
are the over-generating pair, Algeria at 89\% and Mauritania at 77\%. Two cheap choices that each look marginal in isolation, a post-hoc cap worth 1.0 points and a decoder default, are worth 2.3 together on the pooled adapter. Neither is a
model change, and we found the interaction only by testing them jointly.

\section{Statistical Significance}
\label{app:sig}

Every comparison below is a paired bootstrap over utterances with 4{,}000 resamples, the same
test applied to all three subtasks. $P(\text{better})$ is the fraction of resamples in which
the second system has the lower WER, and intervals are percentiles of the resampled
difference. $\Delta$WER is in WER points, and is the mean of the resampled differences rather than the difference of the two point estimates, so it can differ from a subtraction of the WERs printed in the result
tables by a few hundredths of a point.

\paragraph{Subtask 1.3.} Table~\ref{tab:sig} gives the comparisons behind Section~\ref{sec:t13}.

\begin{table}[h]
\centering
\small
\begin{tabular}{lcc}
\toprule
Comparison (validation) & $\Delta$WER & $P(\text{better})$ \\
\midrule
\texttt{q,v} $\rightarrow$ wide+reg & $-2.17$ & 99\% \\
wide $\rightarrow$ wide+reg         & $-0.40$ & 78\% \\
wide+reg $\rightarrow$ soup ($N{=}3$) & $-0.81$ & 97\% \\
wide+reg $\rightarrow$ soup ($N{=}4$) & $\mathbf{-2.26}$ & \textbf{100\%} \\
soup ($N{=}4$) $\rightarrow$ soup ($N{=}6$) & $+0.61$ & 21\% \\
\midrule
same-run pair $\rightarrow$ two-seed pair & $-2.38$ & 100\% \\
cross-recipe pair $\rightarrow$ two-seed pair & $-1.16$ & 99\% \\
soup ($N{=}4$) $\rightarrow$ two-seed pair & $-0.28$ & 87\% \\
\bottomrule
\end{tabular}
\caption{Paired bootstrap over utterances, 4{,}000 resamples. $\Delta$WER is in WER points.
Above the rule, the ablations behind Section~\ref{sec:t13}; below it, the post-evaluation
seed control of Section~\ref{sec:soup}.}
\label{tab:sig}
\end{table}

With 731 development utterances, differences below roughly 1.5 WER points cannot be resolved.
The regularization step we nonetheless kept is \emph{not} significant on its own ($P=78\%$,
95\% interval $[-1.36,+0.68]$), whereas the cross-run soup is unambiguous ($P=100\%$,
$[-3.91,-1.09]$). We report this because system papers often present chains of sub-point
improvements as if each were established. Across the three subtasks this paper reports
thirty-six bootstrap comparisons with no correction for multiplicity. The marginal ones,
Jordan at 87\%, the four-ingredient soup against the two-seed pair at 87\%, and \texttt{turbo}'s
widening at 88\%, should be read as weaker than their numbers suggest.

\paragraph{Subtask 1.1.} Each specialist is tested on its own country's development split,
against the pooled adapter. Seven of the eight are unambiguous; Jordan, the country with the
least headroom, is not (Table~\ref{tab:sig11}).

\begin{table}[h]
\centering
\small
\setlength{\tabcolsep}{4pt}
\begin{tabular}{lrrc}
\toprule
Country & $n$ & $\Delta$WER & $P(\text{better})$ \\
\midrule
Mauritania & 1{,}600 & $-6.53$ & 100\% \\
Yemen      & 1{,}183 & $-3.90$ & 100\% \\
Algeria    &    726 & $-3.12$ & 100\% \\
Morocco    & 1{,}600 & $-2.88$ & 99\% \\
Egypt      & 1{,}600 & $-1.90$ & 99\% \\
UAE        & 1{,}600 & $-1.60$ & 100\% \\
Palestine  &    900 & $-1.05$ & 100\% \\
Jordan     & 1{,}599 & $-0.44$ & 87\% \\
\bottomrule
\end{tabular}
\caption{Subtask 1.1 specialists, in WER points.}
\label{tab:sig11}
\end{table}

\paragraph{Subtask 1.2.} On the 615 held-out utterances, switching base model is decisive
($-3.76$, $P=100\%$, $[-5.02,-2.52]$) and all of \texttt{turbo}'s fine-tuning is too ($-4.52$,
$P=100\%$). Widening \texttt{turbo}'s target set is \emph{not} ($-1.02$, $P=88\%$,
$[-2.96,+0.53]$), which sharpens Section~\ref{sec:t12}. The 1.02 points we contrast with 8 points on 1.1 are themselves inside the noise. The matched-capacity control keeps 2.65 of the base-model gain
($P=100\%$, $[-3.85,-1.39]$) and the extra rank on \texttt{large-v3} is worth 1.11 points ($P=99\%$, $[-2.04,-0.21]$). The member trained on 1.2 data pooled with Subtask~1.1 audio has no significant accuracy edge over the wide \texttt{large-v3} fine-tune ($-0.33$, $P=69\%$), which is consistent with its value lying in decorrelation rather than accuracy.

\section{How Much Transfers (Subtask 1.3)}
\label{app:transfer}

Because we submitted three systems, we can measure how development gains transfer to the blind
test set. The soup's 2.26-point development gain became 0.54 points (24\%), and ROVER's 0.10-point
gain became 0.06 points (60\%). That second ratio should not be read as stable, since the same
step is neutral on the practice test split. Two factors compress the transfer: the blind test
set is easier than our development split (15.09\% vs.\ 19.46\% for the same model, utterances
averaging 3.1\,s against 5.6\,s). The soup was also \emph{selected} on the development set, so
part of its apparent advantage is selection noise. We encourage participants to report this
ratio whenever multiple submissions make it measurable.

\section{Error Analysis by Script (Subtask 1.3)}
\label{app:script}

\begin{table}[h]
\centering
\small
\begin{tabular}{lrrr}
\toprule
Script & Ref.\ words & Share & Error \\
\midrule
Arabic  & 7{,}887 & 66.7\% & 20.2\% \\
Latin (Fr/En) & 3{,}946 & 33.3\% & \textbf{12.5\%} \\
\bottomrule
\end{tabular}
\caption{Word error (substitutions + deletions) by script of the reference word, best single
model, practice test split.}
\label{tab:script}
\end{table}

Table~\ref{tab:script} splits the practice-test reference words by script. The embedded French
and English words, the phenomenon the subtask is named for, are recognized \emph{better} than
the surrounding Tunisian Arabic despite being a third of all tokens. The
practical implication is that effort spent on French-specific lexicons or language-model
biasing is likely wasted, and that Arabic-centric post-correction risks damaging tokens that
are already correct.

Our primary and hamza-normalized scores differ by only 0.27 points, 14.49 against 14.22, indicating that training
on the organizers' normalized targets taught the model the reference orthographic convention
almost exactly. Collapsing hamza in our hypotheses only, rather than on both sides as the
secondary metric does, degrades WER from 17.19\% to 26.98\%, confirming there is no free gain
in orthographic post-processing.

\section{The Metric Noise Floor (Subtask 1.2)}
\label{app:noise}

The references for this subtask are not orthographically self-consistent. After folding hamza,
ta-marbuta and alif-maqsura variants, 844 word types are still written more than one way,
covering 21.9\% of all tokens, often in near-even proportions, so no system can learn a
spelling that is correct both times. Folding these variants on both sides drops the
zero-shot \texttt{turbo} WER on the full 3{,}152-utterance split from 62.2\% to 57.2\%, so roughly 5 WER points of the reported error are unwinnable. We
recommend reporting a folded score alongside the raw one when judging whether a change on this
data is real.

\section{What Did Not Work}
\label{app:neg}

\paragraph{Speech enhancement and bandwidth extension (1.1).} On the 200 noisiest development
utterances, with the same adapter and decode settings and only the audio differing, a CMGAN
\citep{cao2022cmgan} enhancement front-end cost $+5.0$ WER points and VoiceFixer
\citep{liu2021voicefixer} bandwidth extension cost $+18.0$, both
worse in seven of eight countries. VoiceFixer did extend the bandwidth, taking energy above 4\,kHz from under 0.01\% to 1.07\%, but its log-magnitude correlation to the original below
2\,kHz was only 0.63: it rewrote the band that carries the information.

\paragraph{Constrained decoding instead of a length guard (1.1).} Setting \texttt{transformers}' \texttt{no\_repeat\_ngram\_size} to 4, which forbids any repeated 4-gram within a hypothesis, removes runaway generation on the pooled adapter completely: no development hypothesis then exceeds six words per second, so the length guard
has nothing left to cut. It is nonetheless the weaker fix, buying 0.5 country-average WER points against the guard's 1.0, because forbidding repeated 4-grams also blocks legitimate repetition. The two
do not stack, and we submitted the guard.

\paragraph{Noise augmentation (1.1).} In-batch babble and band-limited Gaussian noise at 8 to
25\,dB SNR, over a matched six-epoch run, made no measurable difference to country-average WER,
under 0.2 points either way.

\paragraph{Cross-task transfer (1.2).} Applying the Subtask 1.1 adapter directly to Subtask
1.2 scored worse than the untouched base model. The model trained on that data helped only as an ensemble member, where its disagreement is the point.

\paragraph{Orthographic post-processing (1.2).} Remapping each word to the spelling preferred
in the training portion cost up to 1.4 WER points, and the harm \emph{scaled with model quality}. A fine-tuned model's context-sensitive spelling beats a corpus-wide frequency rule.

\paragraph{More adapter capacity (1.3).} Doubling the LoRA rank to 64 (115.3M trainable)
stalled: held-out loss sat at 0.645 then 0.649 at steps 250 and 500, where the $r=32$ run was
at 0.547 then 0.511 and still falling. The train/held-out gap predicted that regularization
rather than capacity was
the productive direction.

\paragraph{Utterance-level combination (1.3).} Selecting the most central of three hypotheses
per utterance, over the three individually trained models, scored 18.22\%, short of the
17.19\% the soup alone already reached, so word-level voting was the only combination we kept.

\paragraph{Mid-run comparisons across schedules (1.3).} A 5{,}000-step run trailed a
3{,}000-step run on held-out loss at every matched step from 1{,}500 to 3{,}000, and only fell
below the shorter run's best value at step 3{,}750, during the final anneal. Comparing runs of
different total length at matched steps is
misleading.

\section{What the Specialists Learn}
\label{app:whatspec}

Section~\ref{sec:t11} scores each specialist against the pooled adapter on its own country.
That comparison leaves two questions open: whether the gain is specialization or merely more
training, and whether it survives the length guard. Both are settled here on the full
development set, guarded except where stated, so the numbers differ from the unguarded deltas of
Table~\ref{tab:t11}.

\paragraph{Specialization, not extra training.} A specialist continues the pooled adapter on
one country. The matched control continues the same adapter for the same number of optimizer
steps on the \emph{pooled} data, so only the dialect composition of the gradient differs.
Three epochs over 1{,}600 utterances at an effective batch of 16 is 300 steps; one epoch is
100.

\begin{table}[h]
\centering
\small
\begin{tabular}{lrrrr}
\toprule
Country & Steps & Spec. & Control & Spec.\ only \\
\midrule
Algeria & 300 & -3.1 & -0.4 & -2.7 \\
Egypt & 300 & -1.8 & -0.0 & -1.7 \\
Mauritania & 300 & -2.8 & +0.1 & -2.9 \\
Morocco & 300 & -2.2 & +0.6 & -2.8 \\
Jordan & 100 & -0.6 & +0.2 & -0.9 \\
Palestine & 100 & -1.1 & +0.2 & -1.3 \\
UAE & 100 & -1.6 & -0.0 & -1.6 \\
Yemen & 100 & -2.0 & -0.4 & -1.6 \\
\midrule
Mean & & -1.9 & +0.0 & -1.9 \\
\bottomrule
\end{tabular}
\caption{Each specialist against a pooled control given the same number of optimizer steps.
Guarded full-development WER change from the pooled adapter, in WER points. \emph{Spec.\ only} is \emph{Spec.} minus \emph{Control}.}
\label{tab:control}
\end{table}

Neither control moves the pooled adapter (Table~\ref{tab:control}). Three hundred more steps of pooled training change country-average WER by less than 0.1 points and a hundred steps make it 0.2 worse, against the specialists' -1.9. No country gains more than 0.4 points from continued pooled training and Morocco loses 0.6, so the specialist gain is specialization in full, on every dialect.

\begin{table*}[t]
\centering
\small
\begin{tabular}{lrrrrrrrr}
\toprule
Spec. & Egy & Jor & Alg & UAE & Mor & Mau & Yem & Pal \\
\midrule
Egy & \textbf{-1.8} & +2.3 & -0.3 & +1.8 & +2.2 & +0.6 & +0.7 & +2.1 \\
Jor & +2.0 & \textbf{-0.6} & +1.4 & +1.5 & +1.5 & +0.2 & +1.8 & +1.1 \\
Alg & +1.4 & +1.9 & \textbf{-3.1} & +1.0 & +4.0 & +0.6 & +1.3 & +1.8 \\
UAE & +0.8 & +1.4 & -1.1 & \textbf{-1.6} & +1.2 & +0.3 & -0.4 & +1.2 \\
Mor & +1.5 & +2.9 & +1.3 & +1.9 & \textbf{-2.2} & +1.2 & +1.7 & +2.4 \\
Mau & +2.7 & +3.7 & +0.6 & +2.5 & +3.4 & \textbf{-2.8} & +2.5 & +2.0 \\
Yem & +1.1 & +1.6 & +0.9 & +1.3 & +1.2 & +0.7 & \textbf{-2.0} & +1.4 \\
Pal & +1.1 & -0.3 & +0.3 & +0.3 & +1.5 & +0.0 & +0.2 & \textbf{-1.1} \\
\bottomrule
\end{tabular}
\caption{Change in guarded development WER, in WER points, from the pooled adapter when each
specialist (row) decodes each country (column). Bold is the specialist's own dialect.}
\label{tab:xdialect}
\end{table*}

\paragraph{Continuing beats starting over.} Section~\ref{sec:t11} claims the specialist keeps
the shared representation instead of competing with it. Training a fresh adapter on the same
country for the same 300 steps, rather than continuing the pooled one, is worse than not
specializing at all. Mauritania reaches 80.1 and Algeria 74.0, against 76.2 and 69.7 for the pooled adapter itself and 73.3 and 66.6 for the specialists. Sixteen hundred utterances
are not enough to learn a dialect from the base model, and the pooled initialization is what
supplies the rest.

\paragraph{The guard does part of the specialist's work.} Unguarded deltas in this paragraph
come from the same full-development re-scoring as the rest of this appendix, which reproduces
Table~\ref{tab:t11}'s unguarded deltas to within 0.2 points. Under the guard, six of the eight specialist gains move by at most 0.6 points, but Mauritania falls from $-6.3$ to $-2.8$ and Yemen from $-3.9$ to $-2.0$. On the two dialects that over-generate, about
half of what the specialist appeared to buy is runaway generation that the free post-hoc cap
of Section~\ref{sec:t11} already removes. The two interventions are partly redundant, and
exactly on the dialects where specialization looked strongest.

\paragraph{Specialists interfere.} Table~\ref{tab:xdialect} decodes every specialist on every
country.

Every specialist is the best of all the adapters we decoded on its own dialect, yet costs a mean of $+1.4$ points on the seven others against a mean own-dialect gain of $-1.9$. A wrong route
therefore costs about seven tenths of what a right one gains, which is what limits the
front-end of Appendix~\ref{app:routing}.

\section{Routing Without the Dialect Label}
\label{app:routing}

Subtask 1.1 gives the country at test time and Subtask 1.2 gives nothing, so we use 1.1 to
measure what a front-end would recover, precisely because there the routing can be scored.

\paragraph{A linear probe on frozen features.} We mean-pool the frozen
\texttt{whisper-large-v3-turbo} encoder over each utterance's valid frames and fit
multinomial logistic regression on the 12{,}800 training utterances. There is no fine-tuning
and no decoding. It identifies the country of 65.3\% of the 10{,}808 development utterances
against a 12.5\% chance rate, and its confusions are regional rather than arbitrary: the
largest are Algeria with Morocco, then Egypt with Jordan and Jordan with UAE. Utterance ids
share a prefix across splits for 20.8\% of the development set, but accuracy on that
overlapping fifth is \emph{lower} (60.1\%) than on the rest (66.7\%), so the figure is not
inflated by having heard the source.

\paragraph{What routing recovers.} Each utterance is transcribed by the specialist the probe
predicts \emph{for that utterance}. Composing a confusion matrix with
Table~\ref{tab:xdialect} instead would flatter the result, because a router's mistakes fall
on the utterances a model is already likely to get wrong; that approximation gives 51\% where
the exact figure is 44\%.

\begin{table}[h]
\centering
\small
\begin{tabular}{lcc}
\toprule
System & Country-avg WER & $P(\text{better})$ \\
\midrule
pooled adapter          & 56.56 & --- \\
soup of all eight       & 56.31 & 99\% \\
routed by the probe     & 55.73 & 100\% \\
oracle routing          & \textbf{54.66} & 100\% \\
\bottomrule
\end{tabular}
\caption{Subtask 1.1 development, guarded, in \%. $P(\text{better})$ is against the pooled adapter,
paired bootstrap over utterances with 4{,}000 resamples. Scored by the per-utterance routing
pass, which reproduces Appendix~\ref{app:posteval}'s pooled figure to within 0.02 points.}
\label{tab:routing}
\end{table}

Gold routing is worth 1.9 country-average WER points, a 3.4\% relative reduction. That is the most a router can recover, and we call it the prize below. The probe recovers 0.8 points of it, 44\% (Table~\ref{tab:routing}), at $P=100\%$. The gap to the oracle is itself significant ($P=100\%$, 95\% interval $[-1.2,-0.9]$), so the identifier and not the specialists is the binding constraint.
Routing \emph{hurts} on Jordan, 48.7 to 49.5, the country with both the lowest router accuracy, 57.5\%, and the smallest specialist gain.

\paragraph{Under a stronger decoder.} Repeating the whole matrix at beam 5 reproduces the
progression of Appendix~\ref{app:posteval} to within 0.04 points and leaves the conclusion intact.
The oracle prize narrows a little, from 1.9 to 1.8 points, and probe routing recovers 41\% of it
rather than 44\%, still at $P=100\%$. Beam search and specialization are mildly redundant, as
the length guard is, but routing survives both.

\paragraph{The identifier is the improvable part.} Tuning the regularization of the same
linear model, selected on a held-out slice of the \emph{training} split so the development
set stays clean, raises accuracy from 65.3\% to 69.5\% and the recovered fraction from 44\%
to 47\%. Small multilayer perceptrons on the same features do no better, so the ceiling is
the representation rather than the classifier. That four points of accuracy are available
this cheaply, while the gap to the oracle stays significant, is the clearest sign that the
identifier and not the specialists is where further effort belongs. NADI 2026 ran a spoken
dialect identification task alongside the ASR subtasks, and we entered only the ASR ones. On
the evidence above that was a missed opportunity. An identifier built for that track drops
directly into the front-end measured here, and the conversion is already visible. Tuning the same linear model bought four points of identification accuracy and three percentage points of the oracle prize.

\paragraph{Merging is not a substitute.} If the specialists could be combined into one
model the identifier would be unnecessary. Souping all eight into a single rank-256
adapter, by the same exact-mean construction as Section~\ref{sec:soup}, reaches only 56.31, 13\% of the prize and significantly worse than probe routing ($P=100\%$).
Averaging cancels most of the specialization, because each specialist's own-dialect gain
is offset by the seven others' interference on that dialect. The router is a binding
constraint, not an implementation detail.

\end{document}